%% file: main.tex
\documentclass[letterpaper, 10 pt, conference]{ieeeconf}  

\IEEEoverridecommandlockouts                              

\usepackage{hyperref} 
\usepackage{mathrsfs}
\usepackage{amsfonts}
\usepackage{amsmath}
\usepackage[flushleft]{threeparttable}
\usepackage{tablefootnote}
\usepackage{multirow}
\usepackage{hhline}
\usepackage{graphicx}
\usepackage{subfigure}
\usepackage{hanging}
\usepackage{color}
\usepackage{tikz}
\usetikzlibrary{calc,arrows,decorations.markings}
\usepackage{balance}
\usepackage{bbm}
\usepackage{gensymb}
\usepackage{cite}
\usepackage{booktabs}
\usepackage{makecell, booktabs}

\let\labelindent\relax
\usepackage{enumitem}

\graphicspath{{../},{../image}}

\newcommand{\Random}{Agn-Rnd}
\newcommand{\Cascade}{Ret-Gr}
\newcommand{\ObjGrasp}{Obj-Gr}
\newcommand{\CascadeOurs}{CG+Ret}

\newcommand{\noTarget}{\textit{not\_target}}

\newcommand{\Real}{{\mathbb{R}}}

\newcommand{\FCred}{f_{\text red}}

\newcommand{\img}{I}
\newcommand{\cmd}{C}
\newcommand{\graspConfig}{\mathcal{G}}
\newcommand{\grasp}{g}

\newcommand{\featVis}{\mathcal{F}_{\text vis}}

\newcommand{\yVIS}[1][]{y_{\text {vis}{#1}}}
\newcommand{\fVis}{f_{\text{vis}}}

\newcommand{\yProb}{\rho}
\newcommand{\yPos}{r}
\newcommand{\yImgFeat}{y_{\text {img}}}
\newcommand{\dimImg}{{D_I}}

\newcommand{\numProp}{N_p}

\newcommand{\featMerge}{\mathcal{F}_{\text M}}
\newcommand{\BackBone}{ResNet-50}

\newcommand{\wmbd}{e}
\newcommand{\dimEmb}{{D_W}}
\newcommand{\word}{{word}}
\newcommand{\untoken}{{\langle \text{unk} \rangle}}
\newcommand{\dimCmd}{{D_C}}
\newcommand{\fCmd}{f_{\text{cmd}}}
\newcommand{\yCmd}{y_{\text{cmd}}}

\newcommand{\fMerge}{f_{\text M}}
\newcommand{\yMerge}{y_{\text M}}
\newcommand{\cardOrient}{N_{\text{orient}}}
\newcommand{\OutLayer}{f_{\text{grasp}}}
\newcommand{\fGrasp}{\mathcal{F}_{\text{grasp}}}
\newcommand{\yGrasp}[1][]{y_{\text{grasp}{#1}}}
\newcommand{\probGrasp}{\gamma}
\newcommand{\posGrasp}[1][]{\grasp_{0:3{,#1}}}

\newcommand{\algNameFull}{{\em Command Grasping Network}}
\newcommand{\algName}{{\em CGNet}}

\newcommand{\topSpace}{\vspace*{0.06in}}

\title{\LARGE \bf
A Joint Network for Grasp Detection Conditioned on Natural Language Commands}

\author{Yiye Chen$^{1}$, Ruinian Xu$^{1}$, Yunzhi Lin$^{1}$,
        and Patricio A. Vela$^{1}$
\thanks{$^{1}$ Y. Chen, R. Xu, Y. Lin, and P.A. Vela are with the
School of Electrical and Computer Engineering, and the
Institute for Robotics and Intelligent Machines, Georgia Institute of
Technology, Atlanta, GA.  
{\tt\small \{yychen2019, rnx94, ylin466, pvela\}@gatech.edu}}%
\thanks{This work was supported in part by NSF Award \#1605228}%
}

\begin{document}

\maketitle
\thispagestyle{empty}
\pagestyle{empty}


\input{abstract.tex}

\input{intro.tex}

\input{related.tex}
\input{appOverallFig.tex}
\input{relatedNLOR.tex}
\input{relatedGrasp.tex}

\input{probState.tex}

\input{approach.tex}

\input{appVisFeat.tex}

\input{appCommFeat.tex}

\input{appOutLayer.tex}
\input{appLoss.tex}

\input{expIntro.tex}
\input{expDesign.tex}
\input{expData.tex}
\input{expTrain.tex}
\input{expPhy.tex}
\input{expMetrics.tex}

\input{resIntro.tex}
\input{resVision.tex}

\input{resPhysical.tex}

\input{conclusion.tex}


\bibliographystyle{IEEEtran}
\bibliography{main.bib}

\end{document}

%% file: abstract.tex
%
%
\begin{abstract}
We consider the task of grasping a target object based on a natural
language command query. Previous work primarily focused on localizing
the object given the query, which requires a separate grasp detection
module to grasp it.  The cascaded application of two pipelines
incurs errors in overlapping multi-object cases due to ambiguity in
the individal outputs.  This work proposes a model named {\algNameFull}
(\algName) to directly output command satisficing grasps from RGB image
and textual command inputs.  A dataset with ground truth (image,
command, grasps) tuple is generated based on the VMRD dataset to train the
proposed network.  Experimental results on the generated test set show
that {\algName} outperforms a cascaded object-retrieval and grasp
detection baseline by a large margin.  Three physical experiments 
demonstrate the functionality and performance of {\algName}.
\end{abstract}

%% file: intro.tex
\section{Introduction}



The ability to understand natural language instructions, written or spoken, 
is a desirable skill for an intelligent robot, as it allows non-experts
to communicate their demands. 
For manipulation, grasping the target is an indispensable first step in
fulfilling the request.
This paper addresses the problem of object grasp recognition based on a
natural language query.
Two aspects are included in the problem: \textit{what} is the target object required by the command, and \textit{how} to grasp it.
A substantial body of research focuses on the first aspect
\cite{karpathy2014deep,socher2014grounded,Hu_2016_CVPR,
rohrbach2016grounding,nguyen2019object,Nguyen-RSS-20,
hatori2018interactively},
 termed {\em natural language object
retrieval}, but few consider explicitly and/or jointly the second. 

A naive solution for the second is to first locate the target then plan the
grasp in 3D space \cite{Shridhar-RSS-18}. An alternative is to first use
task-agnostic multiple grasp detection, then select a grasp coinciding
with the object using a suitable heuristic.  Both are prone to error in
visual clutter due to imprecise target localization or the inability
disambiguate grasps when objects overlap (see \S\ref{secRes} for
examples).

In this paper, we delay the explicit object retrieval process and
propose a \textit{natural language grasp retrieval} solution that directly
detects grasps based on command understanding as depicted by the
input/output structure of Fig.~\ref{fig:pipeline}.
We argue that doing so is more efficient compared
to cascaded solutions and reduces the aforementioned errors because it
removes the burden of accurate object segmentation, and is less
influenced by distractor objects as they are rarely
included in graspable regions as shown in the output grasp boxes of
Fig.~\ref{fig:pipeline}.
Our hypothesis is that {\em the retrieval and understanding of the target
object can be implicitly done within a deep network, such that 
semantic and object information is encoded into the grasp feature
representation}. 
Confirming the hypothesis involves designing a \textit{natural language
grasp retrieval} deep network to output grasps conditioned on the
input command. The network outputs the grasp box location and the
probability distribution over the classification space of discretized
orientations, retrieval state, and background state.  Two garbage classes
induce decision competition to reject unsuitable candidates.  
Training the network requires the construction of a multi-objects
dataset with associated command and grasp annotations, generated by
semi-automatically augmenting the VMRD dataset
 \cite{roiGrasp2019}
using a templates-parsing approach with paraphrase augmentation for the
command set.
Evaluating trained {\algName} performance on a vision dataset and via physical
experiments validates the hypothesis.


%% file: related.tex
\subsection{Related Work \label{relatedIntro}}

%% file: appOverallFig.tex
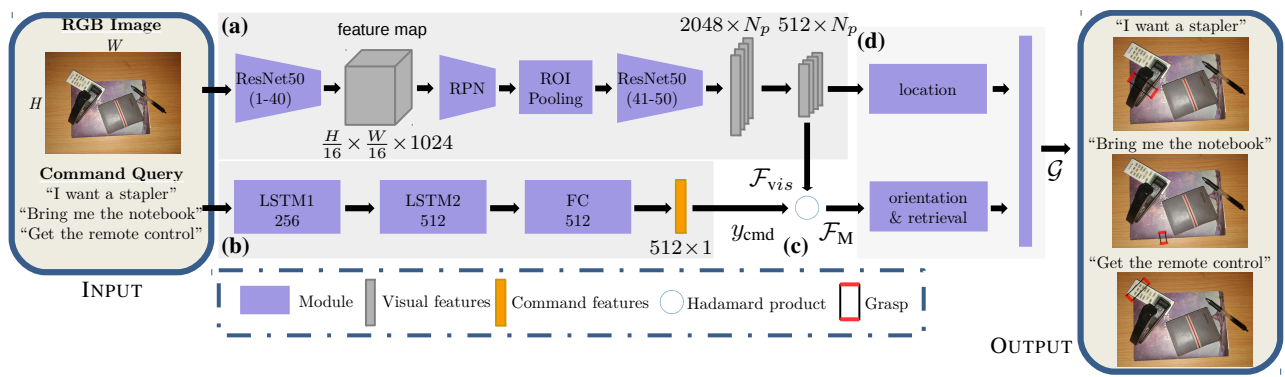
\begin{figure*}[ht!]
  \vspace*{0.06in}
  \centering
  \scalebox{0.97}{
  \begin{tikzpicture}[inner sep = 0pt, outer sep = 0pt]
    \node[anchor=south west] at (0in,0in)
      {{\includegraphics[width=1\textwidth,clip=true,trim=0in
      5.64in 0in 0.035in]{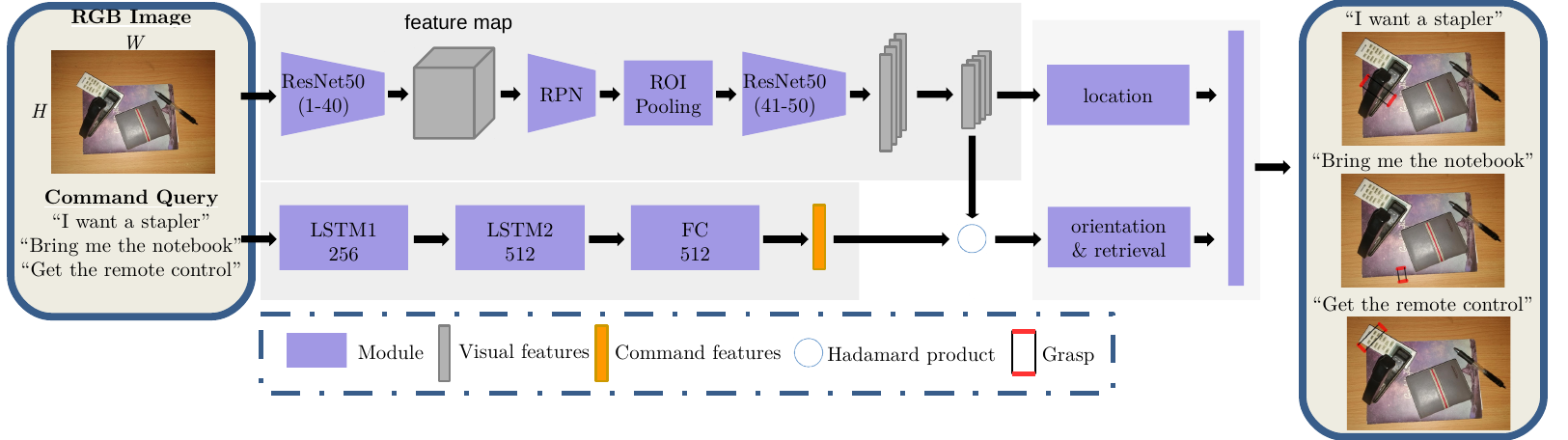}}};
    \node at (4.15in,1.05in) {$\featVis$};
    \node at (4.05in,0.76in) {$\yCmd$};
    \node at (4.5in,0.76in) {$\featMerge$};
    \node at (5.68in,1.10in) {$\graspConfig$};
    \node[anchor=north west] at (1.18in,1.93in) {\small \bf (a)};
    \node[anchor=north west] at (1.18in,0.74in) {\small  \bf(b)};
    \node[anchor=north] at (4.28in,0.74in) {\small \bf (c)};
    \node[anchor=north west] at (4.61in,1.85in) {\small \bf (d)};
    \node[anchor=north] at (0.59in,0.50in) {\small \sc Input};
    \node[anchor=north] at (5.55in,0.20in) {\small \sc Output};
    \node[anchor=south] at (2.07in,1.17in)
    {\footnotesize $\frac{H}{16}\!\times\!\frac{W}{16}\!\times\!1024$};
    \node[anchor=north] at (3.90in,1.93in) 
      {\footnotesize $2048\!\times\!\numProp$};
    \node[anchor=north] at (4.40in,1.93in) 
      {\footnotesize $512\!\times\!\numProp$};
    \node[anchor=south] at (3.66in,0.64in) 
      {\footnotesize $512\!\times\!1$};
  \end{tikzpicture}}%
  \vspace*{-0.1in}
  \caption{Depiction of the deep network processing structure.
    Given an image and a text command: 
    \textbf{(a)} multiple visual grasp features are extracted;
    \textbf{(b)} the command is decoded into a command feature;
    \textbf{(c)} the visual and command outputs are merged;
    \textbf{(d)} and then processed to identify applicable grasps.%
    \label{fig:pipeline}}
  \vspace*{-0.15in}
\end{figure*}

%% file: relatedNLOR.tex
\subsubsection{Natural Langauge Object Retrieval\label{relatedNLOR}}

Natural Language Object Retrieval addresses the problem of locating a
target object in an image when specified by a natural language instruction.
The main challenge lies in interpreting visual and textual input to
capture their correspondence. Early works parsed perceptual and textual
input with human-designed rules, which have less expressive
embeddings \cite{krishnamurthy2013jointly, tellex11}.

The advent of deep learning provided tools to extract high level
semantic meaning from both image and language signals. 
Text-based image retrieval research adopted convolutional neural
networks (CNNs) and recurrent neural networks (RNNs) as image and text 
encoders \cite{socher2014grounded}.
Inspired by the success of object detection, more recent work focuses
on aligning a specific object to the text by segmenting the image and
encoding each object region 
\cite{guadarrama2014open,karpathy2014deep,rohrbach2016grounding,%
nguyen2019object}.
Segmentation modules used include 
external region proposal algorithms \cite{guadarrama2014open, rohrbach2016grounding}, 
object detectors\cite{karpathy2014deep}, or 
built-in object region proposal network \cite{nguyen2019object}. 
Work in \cite{Hu_2016_CVPR} also investigated incorporating both local
and global information for instance-level queries in the text.
The correspondence between visual and textual input can then be
obtained by grounding-via-generation strategy \cite{guadarrama2014open,
Hu_2016_CVPR, rohrbach2016grounding}, 
or designing a proximity-based similarity score function in the shared
image-and-text embedding space such as inner-product \cite{guadarrama2014open, karpathy2014deep} or cosine similarity \cite{Nguyen-RSS-20}. 
Instead of segmenting object-scale regions, our work aims to retrieve
grasp regions.  Though such regions are smaller in scale and contain
object parts, our work demonstrates that sufficient object information
is encoded in the grasp region representations to reason with commands
and differentiate objects.

More specific topics studied in robotics include addressing object attributes specification in commands such as shape \cite{cohen2019grounding}, spatial relations \cite{ Shridhar-RSS-18}, or functionality \cite{Nguyen-RSS-20}.  Sometimes ambiguous commands need clarification through human-robot dialogue \cite{hatori2018interactively, Shridhar-RSS-18}. 
Again, a main difference is the desire to bypass the object retrieval
stage and influence grasp predictions from commands. The research
is complementary.

%% file: relatedGrasp.tex
\subsubsection{Task-agnostic Grasp Detection}
Task-agnostic grasp detection aims to detect all graspable locations
from the perceptual input.
While classical approaches are based on mechanical constrains
\cite{mechanic1, mechanic2}, recent studies resorted to CNN to capture
the geometric information to either predict quality scores for sampled
grasps \cite{Dexnet2}, to directly output the grasp configurations
\cite{lenz2015deep, redmon2015real, watson2017real,kumra2017robotic},
or to capture collision information \cite{murali20206}.
We leverage this research to design a command-influenced grasp
detection network, by augmenting a 2-stage grasp detection network
\cite{chu2018real} whose architecture consists of grasp region 
proposals followed by grasp prediction.

\subsubsection{Semantic Grasp Detection}
Semantic grasp detection seeks functionally suitable grasps for
specific downstream tasks 
\cite{dang2012semantic,rao2018learning,liu2020cage,fang2020learning}, 
such as the
identification of different grasps on an object for different purpose.  
More closely related to our task is that of grasping of a specific
target in a cluttered environment 
\cite{guo2016object,danielczuk2019mechanical,jang2017end,
roiGrasp2019}, where shared encoding of object and grasp content is
shown to be important. Moving to task-agnostic grasp detection
increases the sequential complexity of the solution for object
retrieval \cite{guo2016object,danielczuk2019mechanical}.
As noted earlier, the problem at hand also needs to address textual
comprehension from the input command and its alignment with the grasp
regions. While \cite{roiGrasp2019} investigated reasoning over
object-scale regions, our hypothesis 
allows omission of the object classification in image-text feature fusion.


%% file: probState.tex
\subsection{Problem Statement\label{probState}}

Given an RGB image $\img$ and a corresponding natural language command
$\cmd$ (e.g., \textit{"Give me the banana"}) requesting an object, the {\em command-based grasp selection} problem
is to generate a set of grasp configurations $\graspConfig$ capable of
grasping and retrieving the object if it is present in the image.
It requires establishing a function $f$ such that
$\graspConfig = f(\img, \cmd)$.
The envisioned end-effector is a gripper with two parallel plates, such
that grasping is executed with a vertically downward grasp approach
direction relative to a horizontal surface.  Outputting grasps as a 5D
vector,
$g = [ x, y, \theta, w, h]^T$\!\!\!,\;
defined with respect to the 2D image $\img$ is sufficient to plan such
a grasping action.  The 5D vector $g$ describes the region in the image
plane that the manipulator gripper should project to prior to closing the
parallel plates (or equivalent) on the object.  The coordinates $(x, y,
\theta)$ are the grasp center and orientation (relative to the world
$z$-axis), and $(w, h)$ are the width and height of the pre-closure
grasp region.

The scene may contain a single or multiple objects. The target object can
be partially occluded and the function $f$ should still provide a grasp if
enough of the object is visible and can be grasped.  It should also 
determine whether a detected grasp is suitable or not from the input command.

%% file: approach.tex
\section{Proposed Approach}




The network architecture for grasps recognition based on visual and text
query inputs is depicted in Figure \ref{fig:pipeline}. 
Inspired by the success of feature merging for joint image-and-text
reasoning problems \cite{huk2018multimodal,Kim2017,nguyen2019object}, we
integrate natural language command interpretation with grasp detection
via merged visual and textual representations.  
To facilitate feature merging, a two-stage grasp detection model provides
the base network structure \cite{chu2018real} so that the merger occurs
in the second stage.
Consequently, the pipeline in Figure \ref{fig:pipeline} depicts two
independent and parallel processes first: an image feature extractor to
get a set of visual grasp-and-object sensitive features from the input
RGB image, and a language feature extractor to get the command feature
representation.  
The image feature extractor relies on a region proposal network to
identify a set of potential graspable regions from which the visual
features are obtained.  The final stage fuses both feature spaces and
predicts the {\em 5D parameters} of candidate grasps plus the
{\em suitability} of the grasps for the given command. 
This section covers the details of the network structure. 

%% file: appVisFeat.tex
\subsection{Grasp Region Feature Extraction}

The first stage of the visual processing pipeline proposes a set of grasp
regions of interest (ROIs) \cite{chu2018real,roiGrasp2019}, and their
feature representations, which are expected to contain not only
geometric information for determining grasp configuration, but also
semantic information for reasoning with commands.  
At the end of the grasp proposal pipeline \cite{chu2018real} the
fixed-size feature maps are passed to a shallow convolutional network
to produce a set of vectors in $\Real^\dimImg$, which are interpreted
as the embedding for each candidate grasp region.  The output is set of
such vectors,
\begin{equation} \nonumber
\begin{split} \nonumber
  \featVis & = \{ \yVIS[,i] \}_{1}^{\numProp} 
           = \{(\yProb^i, \yPos^i, \yImgFeat^i)\}_{1}^{N_p} 
           = \fVis (I),
\end{split}
\end{equation}
where the coordinates of $(\yProb, \yPos, \yImgFeat \in \Real^\dimImg ) $
consist of the proposal probability, the predicted position, and the
grasp region feature representation.

A natural approach would be to sequentially apply object detection then
grasp recognition in a {\em detect-then-grasp} pipeline, where the object
of interest comes from the command interpreter. Sequential use with
independent training does not exploit a deep network's ability to learn 
joint representations. Moving to a {\em detect-to-grasp} paradigm
\cite{roiGrasp2019} creates multiple branches after ROI pooling for
learning joint representations similar to the {\em detect-to-caption}
pipeline of \cite{nguyen2019object}. 
In our case the {\em object} category is not given, nor visually
derived, but must be decoded from the command. Since the object
category is still unknown, it should not have primacy. The decision 
should be delayed.  To give grasps primacy, the ROI detector and
feature space primarily key on grasp regions, not object regions.
Filtering grasp candidates by object category occurs when the visual
grasp and command feature spaces merge.  By virtue of the multi-task
objective and joint training, the process of detecting the target object
is implicitly accomplished within the network.  A benefit is that the
grasp ROIs have object-specific attention-like characteristics, which
comes from our observation that a grasp coordinate output $g$ generally
targets a specific object and rarely includes other objects
(see \S\ref{secRes}).
As a
result, it can reduce confusion when there are occlusion cases due to
multiple objects in the scene.  This deep network design reflects a 
{\em grasp-to-detect} paradigm.
The results in \S \ref{secRes} will demonstrate that it is feasible to
encode the object information into these smaller regions, which is
critical for reasoning with commands.

%% file: appCommFeat.tex
\subsection{Command Feature Extraction}
Encoding a natural language command into a vector feature involves first
mapping the command into a sequence of vectors using a trainable word
embedding table. The vector sequence is then passed to a command encoder
to map the sequence to a single vector representation. 

\subsubsection{Word Embedding} A word embedding table is a set of vectors 
$\{\wmbd_i \in \Real^{\dimEmb}\}_{i=1}^n$ representing $n$ words separately, so
that each word $\word_i$ is encoded as $\wmbd_i$. 
The set of $n$ known
words and their vector representations is called the {\em dictionary}.  
For robustness to out-of-dictionary words, an unknown word token 
$\untoken$ is appended to the dictionary. 
The embedding vectors will be learnable so that they can be
optimized to the problem described in \S\ref{probState}.

\subsubsection{Command Encoder} 
The encoder is a 2-layer LSTM \cite{LSTM} with hyperbolic
tangent ($\tanh$) activation. 
LSTM networks extract content or knowledge from sequences, which makes
them effective for joint vision and language problems
\cite{huk2018multimodal,Kim2017}. 
An additional fully connected layer (FC 512) with ReLU activation is
added after the second LSTM block to increase the model's capacity.
The LSTM maps each input vector (the embedding of a word in the command) 
to an output vector that factors in earlier input data due to its feedback
connection.  
After sequentially providing the entire command the last sequentially
output vector defines the command feature representation output. 
The command feature inference is summarized as:
$
  \yCmd = \fCmd(C) \in \Real^{\dimCmd}.
$

%% file: appOutLayer.tex
\subsection{Feature Merging and Output Layer}

The second stage takes the command vector $\yCmd$ and a set of grasp region
features $\featVis$ as input, then predicts grasp configurations and their
matching probability for the command query:
\begin{equation} \nonumber
\fGrasp = \{ \yGrasp[,i] \}_{1}^{\numProp} \!
 	= \{ (\probGrasp_i, g_i) \}_{1}^{\numProp}
 	= \OutLayer (\featVis, \featMerge, \yCmd)
\end{equation}
with $(\probGrasp, g)$ the classification probability
and grasp vector.

The objective is to merge the grasp candidate information with the
command information for conditioning the retrieved grasps in the visual
and textual inputs. 
Prior to merging the two intermediate output streams, a fully-connected
layer $\FCred$ performs dimension reduction of the visual signals in 
$\Real^{\dimImg}$ to match the command signal in $\Real^{\dimCmd}$ when
there is dimension mismatch (e.g., $\dimImg > \dimCmd$).
The Hadamard product (point-wise product) then merges the features: 
\begin{equation} \nonumber
  \yMerge = \fMerge(\yImgFeat,\yCmd) = \FCred(\yImgFeat)\odot \yCmd,
\end{equation}
and get feature set $\featMerge$, where $\odot$ is the element-wise product. 

The final computation involves two sibling, single-layer,
fully-connected networks for grasp fitting, with a position regression
branch and an orientation classification branch \cite{chu2018real}. 
For the position regression branch ({\em location} block), the
position of a grasp should not be conditioned on the command query.
Accordingly, the network receives the image branch outputs
$\featVis$ as inputs, and outputs a 4D position ($x,y,w,h$) for each
orientation class. 
The orientation classification branch (lower block) includes output
classes for rejecting the candidates in $\featVis$ and is where the
command feature influences the outcome.  
The input is a merged feature $\yMerge$.  The output
orientation space consists of $\cardOrient$ classes plus two additional
classes for rejecting grasp candidates that are not sensible. The first,
from \cite{chu2018real}, is a \textit{background (BG)} class.  The second is
a {\noTarget} {\em (NT)} class for feasible grasps that do not reflect the
target object associated to the command request.  The two classes differ
since the \textit{BG} class indicates regions where it is not
possible to grasp (e.g., no object should be there).
Although a double-stream setup (i.e, singling out the language retrieval score
prediction as a separate branch) is also an option
\cite{nguyen2019object, jang2017end}, combining commands and grasps into
the same outcome space (i.e., orientation class) eliminates the need for
a retrieval confidence threshold and employs decision competition to
determine the preferred outcome, 
resulting in a set of candidate grasps $\graspConfig$.



%% file: appLoss.tex
\newcommand{\LTotal}{\mathcal{L}}
\newcommand{\LProp}{{\mathcal{L}}_{\text{p}}}
\newcommand{\LGrp}{{\mathcal{L}}_{\text{g}}}
\newcommand{\LClsFunc}{L_{\text{cls}}}
\newcommand{\LPosFunc}{L_{\text{loc}}}
\newcommand{\Indicator}[1][*]{{\textbf{1}{[#1]}}}

\newcommand{\GTProbProp}{\rho^*}
\newcommand{\GTPosProp}{r^*}
\newcommand{\GTClsGrp}{c^*}
\newcommand{\GTPosGrp}{r^*}

\newcommand{\RPNClsNum}{N_{\text{cls}}^p}
\newcommand{\OutClsNum}{N_{\text{cls}}^g}

\newcommand{\nfGrasp}[1][{}]{Z_{\text{cls}}^{#1}}
\newcommand{\nfLoc}[1][{}]{Z_{\text{loc}}^{#1}}

\subsection{Loss}

Though the network structure includes a second command branch that merges
with the main grasping branch, the loss function for it does not
significantly differ from that of \cite{chu2018real}.  
It consists of proposal $\LProp$ and grasp configuration $\LGrp$
losses, $\LTotal = \LProp + \LGrp$, to propagate corrections back
through both branches during training.  

The proposal loss primarily affects the grasp ROI branch, e.g., $\fVis$. 
A ground truth (GT) binary proposal position and label
{(for positive ROI)} is defined for each ROI ($\GTProbProp$,$\GTPosProp_i$) 
where the GT binary class label is {\tt True} if a grasp is
specified, and {\tt False} if not. 
Both $\GTPosProp$ and $\yPos$ are 4-dimensional vectors specifying center
location and size: $(x, y, w, h)$. The loss is:
\begin{multline} \nonumber
	\LProp(\{\yProb_i, \yPos_i \}_{1}^{\numProp}, 
           \{\GTProbProp_i, \GTPosProp_i\}_{1}^{\numProp})  \\
    = 
      \nfGrasp[-1] \sum_{i=1}^{\RPNClsNum} \LClsFunc(\yProb_i, \GTProbProp_i) 
	  + 
      \nfLoc[-1] \sum_{i=1}^{\numProp} \GTProbProp_i \LPosFunc(\yPos_i, \GTPosProp),
\end{multline}
where 
$\nfGrasp = \RPNClsNum$ and 
$\nfLoc = {\sum_i^{\numProp} \GTProbProp_i }$ are normalization constants,
and 
$\RPNClsNum$ is a hyperparameter specifying the number of ROIs
sampled for the loss calculation. 
The grasp binary label loss $\LClsFunc$ is the cross entropy loss, and
the grasp location loss $\LPosFunc$ is the 
smooth L1 loss \cite{Fast-RCNN}.

The term $\LGrp$ guides the final grasp detection output $\fGrasp$.
GT position and class for each ROI is denoted as $\GTPosGrp_i$ and
$\GTClsGrp_i$ separately. $\GTClsGrp$ is assigned in the following way:
(1) if the ROI is assigned as negative in the proposal stage, then
$\GTClsGrp$ is set to the \textit{BG} class; (2) if the
positive ROI is associated to a non-target ROI, then the $\GTClsGrp$ is
set to \textit{NT} class; (3) otherwise, $\GTClsGrp$ is set to
the corresponding orientation class according to GT orientation angle.
The grasp loss is:
\begin{multline} \nonumber
	\LGrp(
		\{ (\probGrasp_i, \posGrasp[i]) \}_{1}^{\numProp}, 
		 \{\GTClsGrp_i, \GTPosGrp_i\}_{1}^{\numProp} ) = \\
	\frac{1}{\OutClsNum} \sum_{i=1}^{\OutClsNum} \LClsFunc(\probGrasp_i, \GTClsGrp_i)
	+ \frac{1}{ N_{pos}} 
	\sum_{i=1}^{\numProp} \Indicator[\GTClsGrp_i != \textit{bg}]
	\LPosFunc(\posGrasp[i], \GTPosGrp_i),
\end{multline}
where $\OutClsNum$ is a hyperparameter; $N_{pos}=\sum_{i=1}^{\numProp}
\Indicator[\GTClsGrp_i != \textit{bg}]$; $\Indicator$ is the indicator
function that is one when the condition is satisfied and zero otherwise. 

%% file: expIntro.tex
%
%
\section{Methodology}

This section covers the network configuration and training process for
instantiating {\algName}. 
Training will require an annotated dataset compatible with the network's
input-output structure, whose construction is described here.
Since the network will be evaluated as an individual perception module
and integrated into a manipulation pipeline, this section
also details the experiments and their evaluation criteria.



%% file: expDesign.tex
\subsection{Network Structure and Parameters}

The visual processing backbone network is {\BackBone}.  The base layer
and the first three blocks of {\BackBone} are used as the image
encoder, while the last block and the final global average pooling
layer is used to extract vector representations after ROI-pooling.  The
visual features have dimension $\dimImg = 2048$.  For the command
processing pipeline, the word embedding dimension is $D_E = 128$. 
The output dimension for the LSTM layers are set to 
$y_{LSTM1} \in \Real^{256} $, $y_{LSTM2} \in \Real^{512} $.
The dimension of the textual command feature is set to $\dimCmd = 512$. 
The grasp orientation uses $\cardOrient = 19$ classes. 
(w/$180^{\circ}$ symmetry).

%% file: expData.tex
\newcommand{\objSet}{\mathcal{O}}
\newcommand{\obj}{o}
\newcommand{\objPar}{$\langle obj
\rangle$}

\subsection{Annotated Training Data}
Training the proposed network requires a dataset with tuples 
$(\img, \cmd, \graspConfig)$, 
where the grasp configurations in $\graspConfig$ are associated with the
command $\cmd$.  Not aware of such a dataset, we created one by applying
template-based command generation to the multi-objects VMRD
dataset \cite{roiGrasp2019}. 
The VMRD dataset provides labelled objects and grasps, 
$(\img, \objSet, \graspConfig)$, 
where each grasp $\grasp \in \graspConfig$ is associated with an object 
$\obj \in \objSet$ present in $\img$.
We convert the object label to a natural language command by parsing a
randomly chosen template, such as \textit{"Pass me the \objPar "}, from
a pre-defined set \cite{Nguyen-RSS-20}.
This generation is limited to commands with object categories
explicitly stated, and the object class must be in VMRD. 
The network can learn more free-formed commands, like those
demanding a function instead of an object
\cite{Nguyen-RSS-20}.

\subsubsection{Command Augmentation}
The 11 templates adopted from \cite{Nguyen-RSS-20} only include
\textit{subject-verb-object} and 
\textit{verb-object}. 
To enrich the vocabulary size and syntactic diversity, 
we augment the template set
based on the automatic paraphraser. 
We first append 7 templates with different grammatical structures 
(\textit{e.g. "Fetch that {\objPar} for me"}.
Then a group of commands is generated using the initial template sets
and VMRD object labels, one paraphrase is obtained for each command
using the paraphraser, meaningless paraphrases are filtered out
manually, and finally the new template sets are acquired by removing
the object label in the paraphrases.
The above step is repeated 10 times, with 35 commands generated for each. 
The paraphraser used is QuillBot.
In the end, 123 templates
are generated, that differ from 
the initial set, such as \textit{"Grab the {\objPar} and bring it to
me"} and \textit{"The {\objPar} is the one I must have"}.

\subsubsection{VMRD+C Dataset}
The base VMRD dataset consists of 4683 images with over $10^5$ grasps, and
approximately 17000 objects labelled into 31 categories. It is split into
4233 and 450 images for training and testing set respectively. 
For each object in a source image, an $(\img, \objSet, \graspConfig)$ tuple is
generated with the strategy described above.
Ground truth orientation classes for grasps associated to the target object
are set according to their angle, with the rest are labelled {\em NT}.
Furthermore, command requests with target objects not present in the
image are added to the data set for each image. The command is generated
from a random object excluding the ground truth ones present in the image, with all grasps set to {\em NT}. 
The strategy results in 17387(12972 have-target and 4415 no-target
) training data and 1852 (1381 have-target  and 471 no-target
) testing instances.

%% file: expTrain.tex
\subsection{Training And Testing}
\subsubsection{Training Details}
Merging all ROI vision features with textual features slows down
training. 
Applying feature fusion for only positive regions
\cite{nguyen2019object} is inconsistent with the inference procedure,
where background features will also be merged since no ground truth
label is available.  Instead, we sample an equal number of negative
ROIs as positive ones for feature fusion. The rest are retained for
training the visual feature vector output of $\fVis$.  This strategy
improves convergence.

The initial network is {\BackBone} pretrained on ImageNet,
with the other layers and word embeddings randomly initialized. 
Training for the $\untoken$ token, applies random word dropout
with 0.1 probability.
The number of ROIs sampled for loss calculations are
$\RPNClsNum = \OutClsNum = 128$.
The Adam Optimizer\cite{adamOptimizer} with parameters 
$(\alpha, \beta) = (0.9, 0.99)$ and 
initial learning rate of $10^{-4}$ is used.
Training was for $\sim\!400k$ iterations with batch-size of 1. 
\subsubsection{Testing Details}
During testing, 300 ROIs with top proposal scores are sent to the final
grasp prediction layer.  After getting the prediction results, The
higher value from two garbage classes is used as the threshold to
reject candidates independently.  Non-maximum suppression over the
remaining results leads to the output grasp set $\graspConfig$.

%% file: expPhy.tex
\subsection{Physical Experiments}

Three physical experiments are designed to demonstrate the effectiveness and the application value of the proposed method in a perception-execution pipeline.
The objects used are unseen instances of the known categories.

\subsubsection{Single Unseen Object Grasping}
The unseen single object grasping experiment evaluates the generalization
ability of {\algName}.  An unseen instance is randomly placed on the
table, and a command requiring an object type is given. 
For each presented object, we repeated 10 trials for the case where the
required and presented objects match, and 5 trials where there is no match
(i.e., the requested object is not the presented one). 
The command is automatically generated using a random template and an
input object label.  The experiment uses same 8 objects in \cite{roiGrasp2019}.

\subsubsection{Multiple Unseen Objects Grasping}
The aim is to select and grasp the target from amongst multiple
objects, based on the command query.
For each trial, a target object with 4 other interfering objects are
presented, and the target is demanded by command. 
Each target is tested for 10 trials, and cover both cases where the
target is on top or is partially covered.  The target set and the command
generation method is the same as in single-object experiment.

\subsubsection{Voice Command Handover}
A human will submit a command verbally to test {\algName}'s to generalize to 
unknown words or sentence structures.  After the voice command is
translated to text using Google Speech Text API, the experiment follows
the multi-objects experiment. 
The robot arm executes the predicted grasp and passes the object to the
human operator.  The experiment is repeated for 33 trials, with 17
visible targets and 16 partially occluded.

%% file: expMetrics.tex
\subsection{Metrics}

The perception-only experiments test the correctness of the grasp output,
while the manipulation experiments test how well the outputs function in
a perceive-plan-act pipeline with an embodied manipulator. 
The scoring is described below.

\subsubsection{Perception}
We evaluate the model's response to the natural language command
query.  Scoring adopts the object retrieval
\textbf{top-k recall(R@k)} and \textbf{top-k precision(P@k)} metrics to 
evaluate multiple grasp detections \cite{Hu_2016_CVPR}.
R@k is the percentage of cases where at least one of the top-k
detections is correct. 
P@k computes the correct rate for all top-k selections.
A correctly detected grasp has a Jaccard Index (intersection over
union) greater than $0.25$ and an absolute orientation error of less
than $30\degree$ relative to at least one of the ground truth grasps
of the target object.

%

\subsubsection{Physical Experiment} 
To evaluate physical experiment, we separate the pipeline into different
stages and record their success rate respectively.
The three stages considered are: 
\textit{object retrieval}, \textit{grasp detection}, and
\textit{grasp execution}, to identify which step leads to trial
failure.  
The success of the first two are visually confirmed from the
bounding box, while the last requires the target to be grasped, picked
up, and held for 3 seconds.
 The voice command handover experiment
includes the percentage of the spoken command being exactly translated.
The percentage of successful trials, termed \textit{overall} success
rate, is also recorded.

\input{tables/tabVision.tex}

%% file: tables/tabVision.tex
\begin{table}[t!]
  \topSpace
  \caption{VMRD Multi-Grasp Evaluation \label{tab:visionMulti}}
  \centering
  \setlength\tabcolsep{2.0pt}
  \begin{tabular}{ c| c c c c| c c c c |c }
  		\hline
	\textbf{Method} & R@1 & R@3 & R@5 & R@10 & P@1 & P@3 & P@5 &  P@10 & \textbf{FPS}\\ 
 		\hline
 	\Random 		& 25.4 & 25.1 & 26.5 & 24.9 & 27.0 & 26.1 & 25.6 & 25.6 & \textbf{9.1}\\
 	\Cascade 	& 63.1 & 73.1 & 76.5 & 80.8 & 67.4 & 68.5 & 68.3 & 66.5 & 4.5 \\
 	\hline
 	{\algName}           & \textbf{74.9} & \textbf{88.2} & \textbf{91.0} & \textbf{93.2} & 76.1 & 75.3 & 74.8 & 72.2 &  8.7 \\
 	{\CascadeOurs} 	& 74.6 & 86.8 & 88.7 & 90.1 & \textbf{78.0} & \textbf{77.5} & \textbf{76.2} & \textbf{73.3} & 4.4\\
 		\hline
  \end{tabular}
  \vspace*{-0.15in}
\end{table}

%% file: resIntro.tex
\section{Results\label{secRes}}

\input{resVMRDFig.tex}

This section discusses the evaluation results of both the perception
module and the designed physical experiments. Three baselines are
compared with our methods:\\
1) {\em Random (\Random):}
A state-of-art task-agnostic grasp detection network MultiGrasp
\cite{chu2018real} followed by a random choice. 
The model is re-trained on VMRD.\\
2) {\em Cascade (\Cascade):}
A cascading state-of-art natural language object retrieval model \cite{nguyen2019object} and Multigrasp\cite{chu2018real}. 
The target object is the retrieval region with highest score.
Grasps within the retrieval region are kept and ranked based on center-to-center distance (minimum first).
  Both models were re-trained with VMRD+C.
\\
3) {\em \ObjGrasp:}
A grasp detection with object classification network from ROI-based
grasp detection\cite{roiGrasp2019}.  The results are obtained from
their published work, which we take as an upper bound since it skips
the need to interpret commands. 

We also evaluate {{\em \CascadeOurs}}, which takes {\algName} output and
ignores grasps not within the object retrieval region.

%% file: resVMRDFig.tex
\begin{figure*}[t!]
  \topSpace
  \centering
  \scalebox{0.93}{
  \begin{tikzpicture}[inner sep = 0pt, outer sep = 0pt]
    \node[anchor=south west] at (0in,0in)
      {{\includegraphics[width=0.97\textwidth,clip=true,trim=0in 0in 0in 0in]{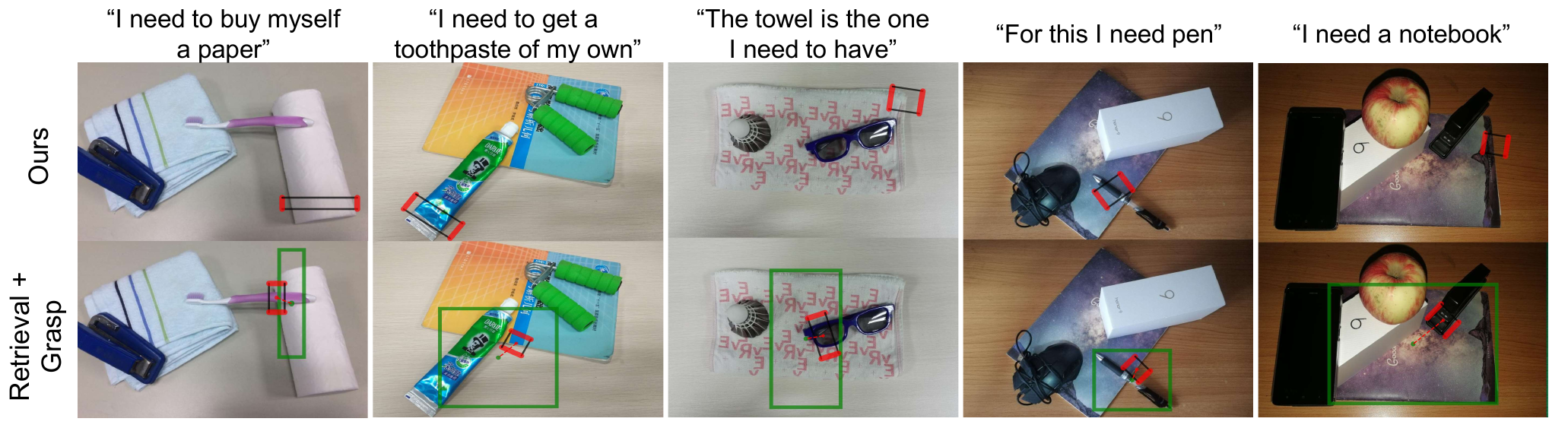}}};
    \node[anchor=south,inner sep=2pt,fill=white] at (0.40in,0.0in) {\small (a)};
    \node[anchor=south,inner sep=2pt,fill=white] at (1.70in,0.0in) {\small (b)};
    \node[anchor=south,inner sep=2pt,fill=white] at (2.97in,0.0in) {\small (c)};
    \node[anchor=south,inner sep=2pt,fill=white] at (4.26in,0.0in) {\small (d)};
    \node[anchor=south,inner sep=2pt,fill=white] at (5.54in,0.0in) {\small (e)};
  \end{tikzpicture}}%
  \vspace*{-0.125in}
  \caption{Grasp detection results for {\algName} and
  {\em \Cascade}. The red box denotes the top
  grasp detection, and the green box object retrieval results. 
  {\algName} avoids mistakes caused by:
    \textbf{(a)(b)(c)} inaccurate localization of the target object;
    and
    \textbf{(d)(e)} target bounding boxes with interfering objects.  %
    \label{fig:resVMRD}}
    \vspace*{-0.15in}
\end{figure*}
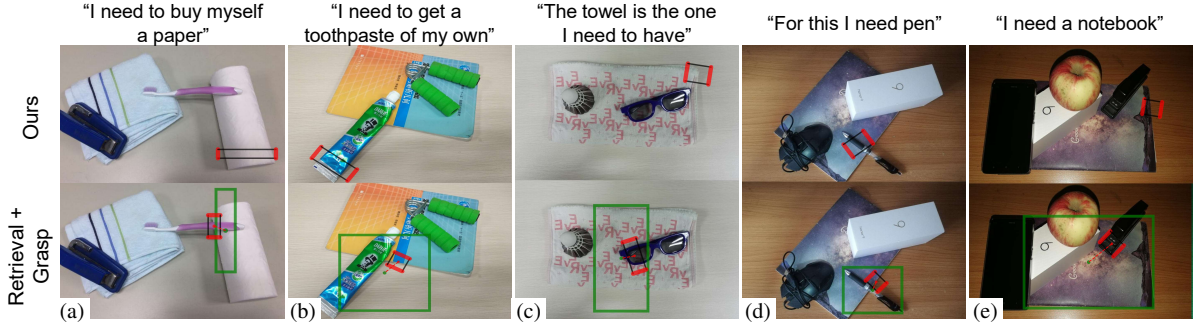

%% file: resVision.tex
\subsection{Vision\label{resVision}} 
Table \ref{tab:visionMulti} lists the visual accuracy and efficiency of
the evaluated methods. All methods outperform {\em \Random}. The hypothesized
value of encoding semantic information into grasp regions and skipping
object detection is evident from the higher values for {\algName} over
{\em \Cascade}, by around 10\% on average.
Examples presented Fig. \ref{fig:resVMRD} demonstrate some of the
problem of the cascade baseline. One class of errors is inaccurate
target localization, where the natural language object retrieval model
yield imprecise locations, which leads to false grasp detection.
{\algName} avoids this step and this type of error. Another class has
errors from overlapping or occluding objects, whereby the object boxes
include distractors objects. Grasp regions are small enough that usualy
asingle object is attached to them, thereby avoiding confusion from
overlapping objects.

\input{tables/TabPhySingle.tex}

{\algName} tests on the 471 \textit{NT} data, for which no grasps should
be proposed, achieves $65.0\%$ success rate, indicating that sometimes
{\algName} fails to recognize objects at the grasp-level.  Applying
the object retrieval information as prior helps distinguish between
objects, as evidenced by the improved P@K of {\em \CascadeOurs} over 
{\algName} alone. The trade-off is a lower recall
thereby causing a drop in R@K value.


%% file: tables/TabPhySingle.tex
\begin{table}[t!]
  \vspace*{-0.040in}
  \caption{Unseen Single Object Grasping \label{tab:phySingle}}
  \vspace*{-0.05in}
  \centering
  \setlength\tabcolsep{1.5pt} 
  \begin{threeparttable}
  \begin{tabular}{|c| c c c | c c c | c c c|}
  		\hline
	\multirow{2}{*}{\textbf{Objects}} &
    \multicolumn{3}{c|}{\textbf{\algName}} &
    \multicolumn{3}{c|}{\textbf{\Cascade}} & \multicolumn{3}{c|}{\textbf{\ObjGrasp \tnote{1}}} \\ 
 		\cline{2-10}
 	{} & Obj & Grp & Exe & Obj & Grp & Exe & Obj & Grasp & Exe \\
 		\hline
 	Apple 			& - & 10 & 9 & 10 & 10 & 8 & - & 10 & 9 \\
 	Banana 			& - & 10 & 10 & 10 & 10 & 10 & - & 10 & 10 \\
 	Wrist Developer & - & 10 & 9 & 10 & 9 & 9 & - & 10 & 10 \\
 	Tape 	   		& - & 10 & 10 & 10 & 10 & 10 & - & 8 & 8 \\
 	Toothpaste 		& - & 10 & 10 & 9 & 9 & 9 & - & 10 & 10 \\
 	Wrench 			& - & 9 & 9 & 5 & 5 & 5 & - & 10& 8 \\
 	Pliers 			& - & 10 & 8 & 1 & 1 & 1 & - & 10 & 10 \\
 	Screwdriver 	& - & 10 & 9 & 5 & 5 & 5 & - & 10 & 9 \\
 		\hline
 	\textbf{Mean} & - & 9.86 & 9.25 & 7.5 & 7.38 & 7.13 & - & 9.75 & 9.25 \\
 		\hline
  \end{tabular}
  
  \begin{tablenotes}\footnotesize
	\item[1] Results adopted from original paper.
  \end{tablenotes}
  \end{threeparttable}
  \vspace*{-0.15in}
\end{table}

%% file: resPhysical.tex
\subsection{Physical Experiments}

\subsubsection{Single Unseen Object Grasping \label{phySingle}}

%
%

The generalization ability of {\algName} is evident in Table
\ref{tab:phySingle}. Though the tested objects were not in the training data
the overall detection and execution success rate matches {\em \ObjGrasp}
which does not have to perform command interpretation.
The \textit{\Cascade} baseline is expected to have strong results also,
but there is a significant performance drop for the \textit{Wrench},
\textit{Pliers}, and \textit{Screwdriver}. It may result from the domain
shift of the unseen objects.

\subsubsection{Multiple Unseen Object Grasping \label{phyMulti}}
Here, all algorithms experience a performance drop as seen in Table 
\ref{tab:phyMulti}.  {\algName} performs closer to {\em \ObjGrasp} than to
{\em \Cascade}, indicating that {\algName} has learnt to encode similar
object-level content in the grasp feature descriptors. However, the
reduced value also indicates that object-level discrimination is not as
strong as it could be. Some form of nonlocal attention is most likely
needed, or loose coupling to object-level feature descriptors.

\input{tables/TabPhyMulti.tex}
\subsubsection{Voice Command Handover}
For the voice command version, the comand interpreter does a better job
than the voice-to-text network, see Table \ref{tab:phyVoice}\;.
Out-of-vocabulary words are input to {\algName} due to translation
error ("want" to "\textit{walked}", "Help" to "\textit{How}")
, unexpected descriptions ("... on the \textit{table}"), 
or colloquial words ("...\textit{please}").
Under these challenges, {\algName} still extracts the key information
and achieves a reasonable $70.0\%(23/33)$ task execution rate. It
outperforms {\em \Cascade} on multiple objects and almost matches its
single object performance. 

All of the outcomes support the value of task prioritization
(here grasps) for contextual interpretation of action commands.
They also indicate that non-local information provides important
information in cases of clutter, where other objects may have similar
grasp-level feature encodings.

\begin{table}[t!]
\vspace*{-0.075in}
  \caption{Voice Command Handover \label{tab:phyVoice}}
  \vspace*{-0.05in}
  \centering
  \setlength\tabcolsep{3.5pt}
  \begin{threeparttable}
  \begin{tabular}{ c| c | c }
  	\textbf{Voice Translation} & \textbf{Grasp Prediction} & \textbf{Execution} \\
  	\hline
  	19/33 & 25/33 & 23/33
  \end{tabular}
 \end{threeparttable}
 \vspace*{-0.15in}
\end{table}

%% file: tables/TabPhyMulti.tex
\begin{table}[t!]
  \vspace*{-0.075in}
  \caption{Unseen Multiple Objects Grasping \label{tab:phyMulti}}
  \vspace*{-0.05in}
  \centering
  \setlength\tabcolsep{1.5pt}
  \begin{threeparttable}
  \begin{tabular}{ |c| c c c | c c c | c c c|}
  		\hline
	\multirow{2}{*}{\textbf{Objects}} &
    \multicolumn{3}{c|}{\textbf{\algName}} &
    \multicolumn{3}{c|}{\textbf{\Cascade}} & \multicolumn{3}{c|}{\textbf{\ObjGrasp\tnote{1}}} \\ 
 		\cline{2-10}
 	{} & Obj & Grp & Exe & Obj & Grp & Exe & Obj & Grasp & Exe \\
 		\hline
 	Apple 			& - & 10 & 8 & 10 & 10 & 10 & - & 10 & 9 \\
 	Banana 			& - & 9 & 9 & 9 & 7 & 7 & - & 10 & 10 \\
 	Wrist Developer & - & 8 & 8 & 9 & 6 & 6 & - & 7 & 6 \\
 	Tape 			& - & 7 & 7 & 9 & 7 & 7 & - & 7 & 7 \\
 	Toothpaste 		& - & 8 & 8 & 5 & 4 & 4 & - & 8 & 8 \\
 	Wrench 			& - & 7 & 6 & 5 & 4 & 4 & - & 10 & 8 \\
 	Pliers 			& - & 7 & 7 & 5 & 2 & 2 & - & 9 & 9 \\
 	Screwdriver 	& - & 7 & 7 & 3 & 3 & 3 & - & 10 & 10 \\
 		\hline
 	\textbf{Mean} & - & 7.88 & 7.50 & 6.88 & 5.38 & 5.38 & - & 8.88 & 8.38 \\
 	\hline
  \end{tabular}
	\begin{tablenotes}
	  \item[1] Results adopted from original paper.
	\end{tablenotes}
	\end{threeparttable}
\end{table}

%% file: conclusion.tex
%
%
\section{Conclusion}
This paper presents \algNameFull (\algName), a network that detects
grasp conditioned on text input corresponding to a natural language command. 
By skipping the object retrieval step and directly detecting grasps, 
{\algName} avoids the errors incurred by inaccurate object localization
and post-processing of cascaded object retrieval and grasp detection
models.  Vision dataset evaluation and three proposed physical
experiment demonstrate the effectiveness and the generalization ability
of {\algName}.

Future work will explore implicit commands where the object is not in the
comand proper but one of its properties is referenced.  We also would like
to incorporate higher-level or nonlocal visual cues to enhance grasp
recognition rates. With both improvements, we envision that the system
would be more effective at interacting with a human.